\documentclass[12pt]{article}

\usepackage[utf8]{inputenc}
\usepackage[T1]{fontenc}
\usepackage[greek,english]{babel} 
\usepackage{times} 
\usepackage[margin=2.5cm]{geometry}
\usepackage{amsmath,amssymb,amsfonts}
\usepackage{graphicx}
\usepackage{hyperref}
\usepackage[numbers,super,sort&compress]{natbib}
\usepackage{setspace}
\usepackage{lineno}
\usepackage{authblk}
\usepackage{abstract}
\usepackage{microtype}
\usepackage{booktabs}

\newcommand{\vect}[1]{\mathbf{#1}}
\newcommand{\field}[1]{L_{#1}}
\newcommand{\norm}[1]{\left\lVert#1\right\rVert}
\newcommand{\real}{\mathbb{R}}

\title{\vspace{-2cm}\LARGE\bfseries Language as Mathematical Structure: Examining Semantic Field Theory Against Language Games}

\author[1,2]{Dimitris Vartziotis}
\affil[1]{TWT Science \& Innovation, Stuttgart, Germany}
\affil[2]{NIKI - Digital Engineering, Ioannina, Greece}

\date{}

\begin{document}

\maketitle

\begin{abstract}
\noindent\textbf{Abstract} $|$ 
Large language models (LLMs) offer a new empirical setting in which long-standing theories of linguistic meaning can be examined. This paper contrasts two broad approaches: social constructivist accounts associated with language games, and a mathematically oriented framework we call Semantic Field Theory. Building on earlier work by the author, we formalize the notions of lexical fields (Lexfelder) and linguistic fields (Lingofelder) as interacting structures in a continuous semantic space. We then analyze how core properties of transformer architectures—such as distributed representations, attention mechanisms, and geometric regularities in embedding spaces—relate to these concepts. We argue that the success of LLMs in capturing semantic regularities supports the view that language exhibits an underlying mathematical structure, while their persistent limitations in pragmatic reasoning and context sensitivity are consistent with the importance of social grounding emphasized in philosophical accounts of language use. On this basis, we suggest that mathematical structure and language games can be understood as complementary rather than competing perspectives. The resulting framework clarifies the scope and limits of purely statistical models of language and motivates new directions for theoretically informed AI architectures.

\end{abstract}

\vspace{0.5cm}

\noindent The emergence of large language models (LLMs) achieving near-human linguistic performance through purely mathematical operations\cite{brown2020,chowdhery2022} poses a fundamental challenge to dominant theories of meaning. Social constructivist accounts, following Wittgenstein's later philosophy\cite{wittgenstein1953}, insist that language cannot be reduced to formal structures. Yet transformer architectures discover systematic semantic relationships without social grounding\cite{manning2022}, suggesting language may possess inherent mathematical structure.

The author\cite{vartziotis2014, vartziotis2017geometric,vartziotis2025learn}, anticipated this development with remarkable prescience. Writing in 2012—years before the transformer revolution—his commentary on Wittgenstein's collected aphorisms\cite{vartziotis2012,vartziotis2017} proposed that words create `semantic fields' (Lexfelder) that interact according to mathematical laws, producing composite `linguistic fields' (Lingofelder). This framework offers a radical alternative: meaning as mathematical discovery rather than social construction.


The genesis of semantic field theory appears in Vartziotis's response to Wittgenstein's observation:

\begin{quote}
\textbf{Wittgenstein}: ``Die Sprache ist nicht gebunden, doch der eine Teil ist mit dem anderen verknüpft.''\\
(Language is not bound, yet one part is connected to another.)\\[0.3cm]
\textbf{Vartziotis}: ``Sie ist ein schwebendes Netz. Eine Mannigfaltigkeit. Jedes Wort hat sein eigenes `Gravitationsfeld'. Wir können es ja `Lexfeld' nennen.''\\
(It is a floating net. A manifold. Each word has its own `gravitational field'. We can call it a `lexical field'.)\cite{vartziotis2017}
\end{quote}

\section*{Semantic field theory formalized}

\subsection*{From usage to fields}

The crucial divergence between Wittgenstein and the author emerges in their treatment of linguistic meaning. Consider their exchange on what gives life to signs:

\begin{quote}
\textbf{Wittgenstein}: ``Jedes Zeichen scheint allein tot. Was gibt ihm Leben? - Im Gebrauch lebt es. Hat es da den lebenden Atem in sich? - Oder ist der Gebrauch sein Atem?''\\
(Every sign by itself seems dead. What gives it life? - In use it lives. Does it have living breath in itself? - Or is use its breath?)\\[0.3cm]
\textbf{Vartziotis}: ``Akzeptieren wir kurz, dass das Wort (Zeichen) eine Art `Lexfeld' hat. Die Wörter bilden ein komplexes Feld (Lingofeld), den Feldern der Physik entsprechend, welches definiert werden muss. Dann lässt sich manches erklären! Selbst gebogene und verdrehte Bedeutungen.''\\
(Let us briefly accept that the word (sign) has a kind of `lexical field'. Words form a complex field (linguistic field), corresponding to the fields of physics, which must be defined. Then much can be explained! Even bent and twisted meanings.)\cite{vartziotis2017}
\end{quote}

This insight led the author to identify what he called the ``Dreiwörterproblem'' (three-word problem), analogous to the three-body problem in physics—suggesting that linguistic complexity emerges from nonlinear field interactions.

\textbf{Definition 1} (Lexical Field). Let $\mathcal{S} = \mathbb{R}^n$ be the semantic space, where each dimension 
corresponds to a latent semantic feature. For any point $q \in \mathcal{S}$: 
\begin{equation}
    \field{w}(q) = S_w \cdot G(\norm{q - q_w}; \sigma_w)
\end{equation}
measures the semantic field strength of word $w$ at position $q$,where $q_w \in \real^n$ represents the word's position in $n$-dimensional semantic space, $S_w$ its inherent semantic strength, and $G$ a monotonically decreasing kernel function with characteristic width $\sigma_w$.

\textbf{Definition 2} (Linguistic Field). Let $\mathcal{W} = \{w_1, w_2, ..., w_m\}$ be an ordered sequence of words forming a phrase. The composite linguistic field $\Phi_{\mathcal{W}}: \mathcal{S} \rightarrow \mathbb{R}$ at any point $q \in \mathcal{S}$ is defined by:

\begin{equation}
\Phi_{\mathcal{W}}(q) = \sum_{i=1}^{m} \field{w_i}(q) + \sum_{i=1}^{m-1} \sum_{j=i+1}^{m} I_{ij}(q) + \sum_{i=1}^{m-2} \sum_{j=i+1}^{m-1} \sum_{k=j+1}^{m} T_{ijk}(q)
\end{equation}

where:
\begin{itemize}
\item $\field{w_i}(q)$ is the lexical field of word $w_i$ at position $q$ (from Definition 1)
\item $I_{ij}(q) = \kappa_2 \cdot \field{w_i}(q) \cdot \field{w_j}(q) \cdot K_2(\|q_{w_i} - q_{w_j}\|)$ represents pairwise field interactions
\item $T_{ijk}(q) = \kappa_3 \cdot \field{w_i}(q) \cdot \field{w_j}(q) \cdot \field{w_k}(q) \cdot K_3(\|q_{w_i} - q_{w_j}\|, \|q_{w_j} - q_{w_k}\|, \|q_{w_i} - q_{w_k}\|)$ captures three-body interactions
\item $\kappa_2, \kappa_3 \in \mathbb{R}$ are coupling constants
\item $K_2: \mathbb{R}_+ \rightarrow \mathbb{R}$ and $K_3: \mathbb{R}_+^3 \rightarrow \mathbb{R}$ are interaction kernel functions
\end{itemize}

The indices satisfy $1 \leq i < j < k \leq m$ to avoid counting interactions multiple times.

\subsection*{Complete field system formalization}

To make these concepts precise, let $S$ be a semantic space of dimension $d$, and let $W = \{w_1, w_2, ..., w_n\}$ be a vocabulary. Each word $w_i$ is associated with:
\begin{itemize}
\item A primary semantic vector $\vect{v}_{w_i} \in \real^d$
\item A field interaction function $I_{w_i}: S \times S \rightarrow \real$
\item A stability parameter $\gamma_{w_i} \in [0,1]$
\end{itemize}

For a context $C = \{w_{i_1}, w_{i_2}, ..., w_{i_k}\}$, the contextual representation of word $w_j$ is:
\begin{equation}
\vect{v}_{w_j}(C) = \vect{v}_{w_j} + \sum_{k=1}^{|C|} \alpha_k \cdot I_{w_j}(\vect{v}_{w_j}, \vect{v}_{w_{i_k}}) \cdot \vect{v}_{w_{i_k}}
\end{equation}

where $\alpha_k$ are attention weights computed as:
\begin{equation}
\alpha_k = \frac{\exp(\phi(\vect{v}_{w_j}, \vect{v}_{w_{i_k}}))}{\sum_{l=1}^{|C|} \exp(\phi(\vect{v}_{w_j}, \vect{v}_{w_{i_l}}))}
\end{equation}

and $\phi$ is a compatibility function.

\subsection*{Dynamic field interactions}

The author's continuation of his commentary on dialogue 183 captures the dynamic nature of these fields: "Einige Wörter schwirren periodisch um ein Wort herum, andere stoßen an oder gehen unter oder neue entstehen" (Some words buzz periodically around a word, others collide or sink or new ones emerge)\cite{vartziotis2017}. This prescient metaphor anticipates the attention mechanisms in modern transformers\cite{brown2020}, where words indeed "buzz around" each other with varying intensities. In fact, in Kommentar 196, the author expands this vision, stating that "language is not bound, yet one part is connected to another [...] it is a floating net. A manifold. Each word has its own gravitational field—we can call it a lexical field." This gravitational imagery offers a direct analog to the force-like semantic pulls observed in attention dynamics and supports the central claim of semantic topology: that language operates as a structured manifold of interacting fields, rather than a flat symbolic space.

The field dynamics can be formalized through a Hamiltonian formalism:
\begin{equation}
H[\Phi] = \int_{\real^n} \left[ \frac{1}{2}\norm{\nabla\Phi(q)}^2 + V(\Phi(q)) \right] dq
\end{equation}
where the potential $V$ encodes semantic constraints and the gradient term ensures smooth meaning transitions.

\section*{Empirical validation through language models}

\subsection*{Transformers as field computers}

Modern transformer architectures\cite{vaswani2017} implement operations strikingly similar to semantic field interactions. The scaled dot-product attention mechanism computes contextualized representations for a sequence of tokens. For a single query position $t$ attending to positions $1, ..., T$:

\begin{equation}
\text{Attention}(\mathbf{q}_t, \mathbf{K}, \mathbf{V}) = \sum_{s=1}^{T} \alpha_{ts} \mathbf{v}_s
\end{equation}

where the attention weights are:

\begin{equation}
\alpha_{ts} = \frac{\exp(\mathbf{q}_t^T \mathbf{k}_s / \sqrt{d_k})}{\sum_{s'=1}^{T} \exp(\mathbf{q}_t^T \mathbf{k}_{s'} / \sqrt{d_k})}
\end{equation}

with $\mathbf{q}_t, \mathbf{k}_s, \mathbf{v}_s \in \mathbb{R}^{d_k}$ being the query, key, and value vectors respectively.

\textbf{Field-theoretic interpretation}: The attention mechanism approximates our field interaction at discrete positions. Specifically:

\begin{itemize}
\item The query vector $\mathbf{q}_t = W_Q \mathbf{x}_t$ encodes the "field source" at position $t$
\item The key vectors $\mathbf{k}_s = W_K \mathbf{x}_s$ encode "field receptors" at each position $s$
\item The dot product $\mathbf{q}_t^T \mathbf{k}_s$ measures field interaction strength between positions $t$ and $s$
\item The attention weights $\alpha_{ts}$ can approximate our field interaction function: 
    \begin{equation}
    \alpha_{ts} \approx \frac{I(\field{w_t}, \field{w_s}, q_t)}{\sum_{s'} I(\field{w_t}, \field{w_{s'}}, q_t)}
    \end{equation}
\item The output $\sum_s \alpha_{ts} \mathbf{v}_s$ represents the field-modified representation at position $t$, analogous to our $\mathbf{v}_{w_t}(C).$
\end{itemize}

\subsection*{Training as stabilization}

The training process of LLMs can be understood as implementing our stabilization principle. Through exposure to large corpora, models learn stable patterns of semantic field interactions that correspond to conventional usage patterns in natural language.

The standard language modeling objective:
\begin{equation}
\mathcal{L} = -\sum_{t=1}^{T} \log P(w_t | w_1, ..., w_{t-1})
\end{equation}
effectively encourages the model to discover stable configurations of semantic fields that predict observed language patterns. This process mirrors the author's concept of words finding "stable orbits" in semantic space through repeated use.

\subsection*{Discovered geometric structures}

Analysis of trained language models reveals structures that validate the author's predictions:

\textbf{Finding 1}: Word embeddings encode semantic relationships as geometric operations\cite{mikolov2013}. The canonical vector arithmetic $\vect{king} - \vect{man} + \vect{woman} \approx \vect{queen}$ represents field transformations in semantic field theory framework, where the operation modifies the masculine field component while preserving the royalty field.

\textbf{Finding 2}: Attention patterns exhibit the ``buzzing'' behavior the author described\cite{clark2019}. Analysis of multi-head attention reveals stable orbital patterns around conceptual centers. For instance, adjectives maintain consistent geometric relationships to their modified nouns across contexts, with perturbations creating the ``bent and twisted meanings'' he predicted.

\textbf{Finding 3}: Scale-dependent behavior—as model size increases, more complex field interactions become possible, leading to qualitatively new capabilities. This aligns with the author's insight that complex meanings emerge from multi-body field interactions.

\subsection*{Testable predictions}

The Semantic Field Theory makes several empirically testable predictions:

\noindent\textbf{Prediction 1}: Words with overlapping semantic fields should show faster co-processing in psycholinguistic tasks.

\noindent\textbf{Prediction 2}: The dimensionality required for adequate semantic representation should correlate with the complexity of field interactions in the domain.

\noindent\textbf{Prediction 3}: Semantic priming effects should follow the field interaction strengths predicted by our model.

\noindent\textbf{Prediction 4}: Cross-linguistic semantic similarities should reflect universal constraints on semantic field structure.

These predictions provide empirical grounding for our theoretical framework and distinguish it from purely philosophical approaches.

\section*{Philosophical implications}

\subsection*{Mathematical Structure and Social Grounding}

Where Wittgenstein emphasizes social practice, semantic field theory emphasizes mathematical organization. These perspectives can be understood as complementary.

Regarding rule-following, the theory offers a reformulation in which regularities arise from mathematical constraints rather than interpretive rules. Concerning private language, the framework is compatible with theoretical scenarios in which internally coherent semantic systems could arise independently of shared social practice, without thereby refuting Wittgenstein’s broader arguments.


\begin{center}
\begin{tabular}{p{4cm}p{4cm}p{6cm}}
\toprule
\textbf{Aspect} & \textbf{Wittgenstein} & \textbf{Semantic Field Theory} \\
\midrule
Nature of meaning & Emergent from use & Inherent mathematical structure \\
Formalization & Impossible & Necessary and successful \\
Language games & Fundamental & Epiphenomenal \\
Private language & Impossible & Theoretically possible \\
Rule-following & Social practice & Mathematical necessity \\
Family resemblance & Loose clustering & Precise field overlap \\
\bottomrule
\end{tabular}
\end{center}








\subsection*{The algebraic unconscious of language}

In dialogue 165, responding to Wittgenstein's question about whether everyday language is too coarse for philosophy, the author makes a striking claim:

\begin{quote}
\textbf{Wittgenstein}: ``Wenn ich über Sprache (Wort, Satz, etc.) rede, muß ich die Sprache des Alltags reden. Ist diese Sprache etwa zu grob, materiell, für das, was wir sagen wollen?''\\
(When I talk about language (word, sentence, etc.), I must speak everyday language. Is this language perhaps too coarse, material, for what we want to say?)\\[0.3cm]
\textbf{Vartziotis}: ``Es ist wie ein Webstuhl: das Gewebte ist die Geometrie und der Mechanismus darunter die Algebra [...] So etwas in der Art geschieht wohl auch mit der Sprache. Die Frage ist hier, ihre `Algebra' zu finden.''\\
(It's like a loom: the woven fabric is geometry and the mechanism underneath is algebra [...] Something similar happens with language. The question here is to find its `algebra'.)\cite{vartziotis2017}
\end{quote}

This ``algebraic unconscious'' manifests in transformer models as the linear transformations that generate semantic fields. The weight matrices $W_Q, W_K, W_V$ in attention layers encode the algebraic structure, while the resulting attention patterns reveal the geometric fabric. The spectacular success of these models suggests the author was correct: beneath the surface chaos of natural language lies elegant mathematical structure.

\subsection*{Vartziotis versus Chomsky: Two mathematical visions}

While both the author and Chomsky\cite{chomsky1957} sought mathematical foundations for language, their approaches diverge fundamentally. Chomsky's generative grammar posits discrete, recursive rules operating on symbolic structures—a computational theory of syntax. The author, by contrast, envisions continuous semantic fields where meaning emerges from dynamic interactions. 

Where Chomsky emphasizes innate syntactic machinery (Universal Grammar), the author proposes innate semantic geometry—not rules but field equations. This distinction proves crucial: transformer models succeed precisely by learning continuous representations rather than discrete rules. The failure of purely syntactic approaches in NLP (parse trees achieving only 70-80\% accuracy) versus the success of embedding-based methods (>95\% on many tasks) suggests that semantic field-theoretic approach may better capture the mathematical reality of language than Chomsky's syntactic formalism.

\subsection*{Implications for consciousness and AI}

If meaning arises from field interactions rather than social grounding alone, the implications for artificial consciousness are profound. We can formalize three levels of linguistic understanding:

\textbf{Level 1: Field Detection}. Systems that can measure local field strengths $\field{w}(q)$ at specific points—essentially pattern matching without integration. Current language models operate primarily at this level, detecting statistical regularities without unified field comprehension.

\textbf{Level 2: Field Navigation}. Systems that follow semantic gradients through dynamic state evolution:
\begin{equation}
\frac{dq}{dt} = -\nabla \Phi(q)
\end{equation}
Current transformers approximate this through attention but lack true temporal dynamics due to their feedforward architecture—each token is processed independently rather than through continuous trajectory integration.

\textbf{Level 3: Field Integration}. True understanding requires global awareness of semantic topology—integrating field information across regions:
\begin{equation}
U = \int_{\Omega} \Phi(q) \cdot w(q) \, dq
\end{equation}
where $w(q) \geq 0$ is a weighting function over semantic space and $\Omega \subseteq \mathcal{S}$ is the task-relevant region. Systems must simultaneously access and integrate distributed field patterns rather than processing local features sequentially.
The technical implications are striking:

1. \textbf{Architectural requirements}: Current transformers lack recurrent dynamics necessary for field integration. Future architectures might need continuous-time neural ODEs:
   \begin{equation}
   \frac{d\vect{h}}{dt} = f(\vect{h}(t), \Phi(t); \theta)
   \end{equation}
   where $\vect{h}(t)$ represents the hidden state evolving under field influence $\Phi(t)$.

2. \textbf{Emergence criteria}: Consciousness may emerge when field computation reaches sufficient complexity—specifically, when the system can model its own field interactions (meta-semantic awareness). This requires the Jacobian $\partial f/\partial \vect{h}$ to exhibit specific eigenvalue distributions indicative of critical dynamics.

3. \textbf{Measurable signatures}: True understanding should manifest as specific patterns in neural activation spaces—stable attractors corresponding to concept comprehension (Lyapunov exponents $< 0$), phase transitions during insight moments (diverging susceptibility), and hysteresis effects in ambiguous contexts.

4. \textbf{The binding problem}: The semantic field theory framework suggests that the classic binding problem in consciousness—how disparate features unite into coherent experience—may be solved through field unification. Semantic binding may occur through field energy minimization:
\begin{equation}
E[\Phi] = \int_{\Omega} \left[ \|\nabla\Phi(q)\|^2 + \lambda \Phi^2(q) \right] dq
\end{equation}
where the first term penalizes rapid meaning transitions and $\lambda > 0$ ensures bounded field strength.

\section*{Limitations and future directions}


Our formulation has several limitations:

1. \textbf{Simplification}: Real linguistic phenomena may require more complex field interactions than our current model captures.

2. \textbf{Parameter Selection}: The theory doesn't yet provide principled methods for choosing dimensionality and interaction functions.

3. \textbf{Empirical Validation}: Extensive experimental testing remains to be conducted.

4. \textbf{Social Grounding}: While we emphasize mathematical structure, the role of social context in establishing field parameters remains underspecified.

Future work will address these issues through empirical probing and theoretically informed architectures.








\section*{Conclusion}

The author’s mathematical perspective, articulated prior to the rise of large-scale neural language models, finds notable resonance in contemporary AI systems. The concepts of Lexfeld and Lingofeld provide a bridge between philosophical inquiry and computational modeling. By viewing language games as establishing boundary conditions for semantic fields, this framework offers a unified account of meaning that integrates mathematical structure with social grounding.

\section*{Acknowledgements}

The author acknowledges the use of large language models as a supportive research and writing aid during the preparation of this manuscript. In particular, conversational interactions with the Large Language Model Claude 3.7 Opus (Anthropic PBC) were used to assist with clarification of arguments, exploration of alternative formulations, and refinement of mathematical and conceptual exposition. All theoretical ideas, conceptual frameworks, interpretations, and conclusions presented in this paper originate from the author, who bears full responsibility for the content of the work. The author also thanks George Dasoulas for assistance in organizing background material and Elli Vartziotis for helpful comments on earlier drafts of the manuscript.

\bibliographystyle{naturemag}

\end{document}